%% file: 00_main_ijrr.tex
\documentclass[Afour,sageh,times]{sagej}
\usepackage{moreverb,url}
\usepackage[colorlinks,bookmarksopen,bookmarksnumbered,citecolor=red,urlcolor=red]{hyperref}
\newcommand\BibTeX{{\rmfamily B\kern-.05em \textsc{i\kern-.025em b}\kern-.08em
T\kern-.1667em\lower.7ex\hbox{E}\kern-.125emX}}
\def\volumeyear{2016}

\usepackage{fancyhdr}

\usepackage{textcomp}\usepackage{gensymb}

\usepackage{multicol}

\usepackage{subfig}
\usepackage{graphicx}

\usepackage{xr}
\usepackage{xr-hyper}

\usepackage[nolist]{acronym}
\usepackage{nicefrac}
\usepackage{upgreek}
\usepackage{units}
\usepackage{afterpage}
\usepackage{booktabs}
\usepackage[binary-units]{siunitx}
\usepackage{tablefootnote}
\usepackage[final]{microtype}
\newcommand{\norm}[1]{\left\lVert#1\right\rVert}
\newcommand{\mat}[1]{{\ensuremath{{\mathbf{#1}}}}}
\newcommand{\gs}[1]{{\ensuremath{{\mathbf{#1}}}}}

\newcommand\footnoteref[1]{\protected@xdef\@thefnmark{\ref{#1}}\@footnotemark}

\newcommand*\m{\,\si{\metre}}

\DeclareSIUnit[per-mode=symbol,per-symbol=p]{\MBps}{\mega\byte\per\second}
\DeclareSIUnit[per-mode=symbol,per-symbol=p]{\Mbps}{\mega\bit\per\second}
\DeclareSIUnit[per-mode=symbol,per-symbol=p]{\MB}{\mega\byte}

\begin{document}

\runninghead{Bernreiter et al.}

\title{A Framework for Collaborative Multi-Robot Mapping using Spectral Graph Wavelets}

\author{Lukas Bernreiter\affilnum{1}, Shehryar Khattak\affilnum{2}, Lionel Ott\affilnum{1}, Roland Siegwart\affilnum{1}, Marco Hutter\affilnum{2} and Cesar Cadena\affilnum{1}}

\affiliation{\affilnum{1}Autonomous Systems Lab, ETH Zurich, Switzerland\\
\affilnum{2}Robotics Systems Lab, ETH Zurich, Switzerland\\}

\corrauth{Lukas Bernreiter, 
Autonomous Systems Lab,
ETH Zurich,
Zurich, Switzerland 8092.}

\email{berlukas@ethz.ch}

\begin{abstract}
The exploration of large-scale unknown environments can benefit from the deployment of multiple robots for collaborative mapping.
Each robot explores a section of the  environment and communicates onboard pose estimates and maps to a central server to build an optimized global multi-robot map.
Naturally, inconsistencies can arise between onboard and server estimates due to onboard odometry drift, failures, or degeneracies.
The mapping server can correct and overcome such failure cases using computationally expensive operations such as inter-robot loop closure detection and multi-modal mapping.
However, the individual robots do not benefit from the collaborative map if the mapping server provides no feedback.
Although server updates from the multi-robot map can greatly alleviate the robotic mission strategically, most existing work lacks them, due to their associated computational and bandwidth-related costs.
Motivated by this challenge, this paper proposes a novel collaborative mapping framework that enables global mapping consistency among robots and the mapping server. 
In particular, we propose graph spectral analysis, at different spatial scales, to detect structural differences between robot and server graphs, and to generate necessary constraints for the individual robot pose graphs.
Our approach specifically finds the nodes that correspond to the drift's origin rather than the nodes where the error becomes too large.
We thoroughly analyze and validate our proposed framework using several real-world multi-robot field deployments where we show improvements of the onboard system up to 90\% and can recover the onboard estimation from localization failures and even from the degeneracies within its estimation.
\end{abstract}

\keywords{Multi-Robot Mapping, Spectral Graph Theory}

\maketitle

\input{01_introduction}
\input{02_related_work}
\input{03_preliminaries}
\input{04_materials_and_methods}
\input{05_experiments}
\input{06_conclusions}

\begin{acks}
This work was supported as a part of NCCR Robotics, a National Centre of Competence in Research, funded by the Swiss National Science Foundation (grant number 51NF40\_185543).

The authors are thankful to Marco Tranzatto, Patrick Pfreundschuh, Samuel Zimmermann and Timon Homberger, Gabriel Waibel for their assistance with field experiments.
\end{acks}

\bibliographystyle{SageH}
\bibliography{bib/cdpgo.bib, bib/gsp.bib, bib/misc.bib, bib/cdpgov2.bib}

\end{document}

%% file: 01_introduction.tex
\section{Introduction}
\label{sec:introduction}
Over recent years, an abundance of localization and mapping frameworks have been proposed and successfully deployed in various robotic scenarios. As part of this research, many traditional SLAM challenges have been fully or partially addressed. 
Despite this development, new challenges readily arise with the need for more robotic autonomy and the deployment of heterogeneous robotic teams in large-scale environments.
In particular, an increase in the number of deployed robots and autonomy requires a higher degree of robustness and efficiency.
At the same time, scalability and persistence across all systems become a pertaining issue. 
While it is difficult to maintain a consistent estimate of the environment across all employed systems, it is an essential prerequisite for operating robotic teams in applications like disaster response or search and rescue. 

With the recent advent of high-bandwidth mobile networks such as 5G networks, collaborative robotic approaches have received increased attention in the robotics community due to their improved practical feasibility.
A promising research direction is to employ a centralized mapping approach. 
Centralized servers running in the local network or a remote cloud environment have more computational capacity than individual robots.
Therefore, they can perform expensive operations such as global optimizations, loop closing, and exploitation of all available sensor data to improve accuracy and overcome onboard failures.

Most collaborative mapping approaches focus on building accurate maps on the server and ignore the use of global multi-robot information to provide localization corrections to individual robots.
Especially in centralized settings without feedback, pose estimation discrepancies may arise between robots during large missions leading to severe drift between robot and server maps resulting in increased optimization time at the server for collaborative mapping. 
Therefore, it is desirable to provide additional constraints to improve onboard estimation and collaborative mapping performance for large-scale multi-robot missions.

Furthermore, multi-robot missions often deploy a heterogeneous set of robots, \textit{e.g.}, aerial and ground robots, which additionally might rely on heterogeneous sensory systems. 
Carrying a diverse set of multi-modal sensors onboard and effectively utilizing different algorithms for, \textit{e.g.}, localization and mapping, can be highly beneficial for the deployment as it becomes more flexible and robust.
However, no common layer sharing data to improve pose estimation and mapping estimates among the employed robots is readily available with diverse sensory systems.
Hence, a sensor modality-invariant approach that can incorporate and communicate relevant consistency information among robots while maintaining low network bandwidth requirements is essential for large-scale multi-robot field deployments.

This paper proposes a novel multi-robot pose graph consistency approach independent of the underlying robot pose estimation processes. 
Our proposed approach relies only on a sparse abstraction of the estimated poses in $SE(3)$.
Moreover, the framework operates in the graph spectral domain of the pose graphs to identify structural anomalies in the individual robot pose graphs using a multi-scale analysis.
By examining the structural components of the pose graphs at different scales, our system identifies discrepancies in the local and coarser neighborhoods and adds corresponding constraints to improve the pose estimation accuracy of individual robots and make the individual robot and collaborative server maps consistent with each other.
The key contributions of this paper are:
\begin{itemize}
    \item Graph spectral analysis of pose graphs to identify discrepancies between onboard and server pose graphs.
    \item Automatic adaptive inference of multi-scale constraints to correct onboard estimation failures.
    \item Comparison against current state-of-the-art approaches on datasets and a thorough quantitative analysis on large-scale multi-robot field deployments are presented to validate the proposed approach.
\end{itemize}
\begin{figure*}[!t]
  \centering
   \includegraphics[width=1.0\textwidth, trim={0.0cm, 0.1cm, 0.0cm, 0cm}, clip]{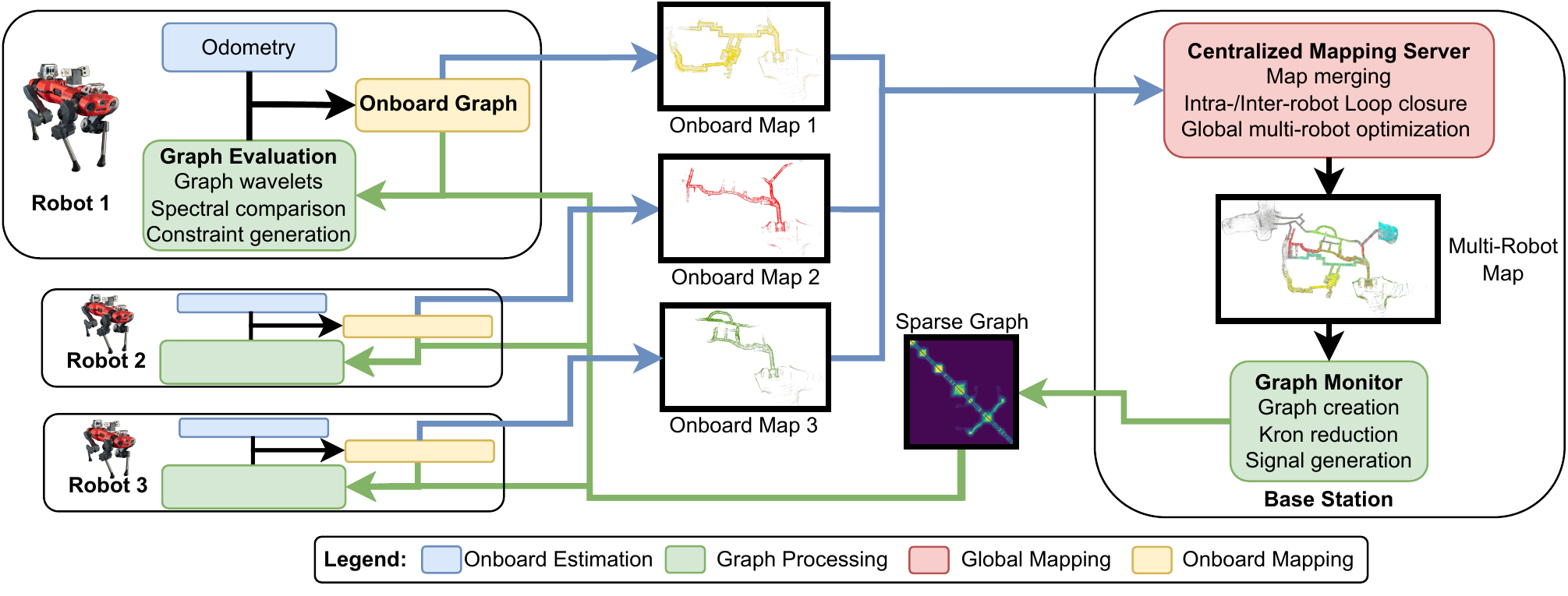}
   \caption{Overview of our approach. We consider multiple robots simultaneously exploring an environment and sending incremental maps to a centralized mapping server. The server accumulates all robot maps and jointly optimizes them. A relaxation of the collaborative multi-robot map is sent back to the robots, where a multi-scale graph spectral analysis is performed to identify discrepancies onboard and server maps and to generate necessary constraints for making them consistent.}
   \label{pics:fgspv2:overview}
\end{figure*}

%% file: 02_related_work.tex
\section{Related Work}
\label{sec:related_work}
In this section, we review the state-of-the-art collaborative multi-robot localization and mapping approaches as well as the current applications of graph signal processing and degeneracy and failure detection.
\subsection{Collaborative Multi-Robot Mapping}
Collaborative multi-robot approaches can be distinguished into centralized~\citep{Deutsch2016a,Schmuck2019,Karrer2018} and distributed solutions~\citep{Cunningham2013,Dong2015}.
\citet{Deutsch2016a} proposed a vision-based centralized multi-robot SLAM approach where a mapping server performs loop closures and replaces robot pose graphs with corrected graphs. 
A similar approach was proposed by~\citet{Schmuck2019} in which robots send local maps to a mapping server which then returns optimized keyframes and landmarks to each robot to include in their onboard optimizations, thus increasing the bandwidth requirements for real-world robot deployments.
The work of \citet{van2018collaborative} proposes an encoding and decoding of visual features during the transmission of the maps to reduce the required bandwidth. 
CoSLAM~\citep{Zou2013} proposes to make use of GPU computing to circumvent the need for large computational processes and improve the speed of onboard optimizing tasks, hence requiring a GPU onboard individual robots.

Different from vision-only approaches, LAMP~\citep{Ebadi2020, chang2022lamp} proposes a large-scale collaborative multi-modal SLAM framework.
However, their proposed approach does not provide any pose corrections from the centralized server to the individual robots.

In contrast to centralized approaches, distributed approaches require each robot to run a full onboard SLAM solution~\citep{Dong2015} and share marginalized information with other robots~\citep{Cunningham2013}, thus making full information available to each robot. 
Additionally, they have the advantage of scaling well to large swarms of robotic systems~\citep{ziegler2021distributed} but typically increase the onboard compute requirements significantly.

A crucial aspect of multi-robot SLAM is the ability to incorporate inter-robot loop closures. 
\citet{BeenKim2010} aims to achieve consistent maps across multiple robots independently of the employed sensing modalities by detecting loop closures between robots and connecting their pose graphs.
In the same direction, ~\citet{Mangelson2018a,Mangelson2019} aim to robustly select inter-robot loop closure candidates by maintaining pair-wise consistent measurements. 
More recently~\citet{Lajoie2020} proposed a distributed system with distributed loop closure detection. 

The more robots are deployed for a specific task, the more information needs to be processed, potentially leading to delays or longer processing times, especially for components such as the factor graph optimization. 
Recently, COVINS~\citep{Schmuck2021} demonstrated a collaborative deployment of 12 individual agents while maintaining a reasonable collaborative trajectory error.
Although their system propagates optimized poses from the centralized server back to individual agents, the poses are only used for drift quantization by comparing the optimized to the onboard estimate. 
Thus, the onboard pose estimations are not corrected.

Concluding, many existing approaches are limited to a single modality only~\citep{Lajoie2020, Karrer2018, Deutsch2016a} often incorporated in tightly coupled multi-robot frameworks, exchanging large data structures such as descriptors~\citep{Tian2022}, partial or complete~\citep{Schmuck2019} factor graphs. 
As a consequence, the systems become less flexible and maintain little versatility for the application of different robotic tasks. 
Conversely, this paper proposes to detect discrepancies between the robot graphs using spectral analysis and a sparse abstraction of the server graph to generate an individual set of constraints for each robot.
Hence, the proposed approach achieves high accuracy and mapping consistency while maintaining low network and compute requirements.
\subsection{Failure and Degeneracy Detection} 
Pose estimation from onboard sensors is subject to drift (accumulation of small errors) and to degeneracies (errors due to specific sensor modality's deficiency). 
Recognizing such errors enables corrective actions to avoid possible catastrophic losses (\textit{e.g.,} platform crashes and wrong decision making).
However, evaluating the quality of poses or maps is not trivial when no ground truth is available for comparison.
In~\citet{Schwertfeger2013}, a metric to assess the quality of the maps was proposed by matching topological graphs from the robot with a ground truth map. 
Some research also approaches the problem using redundant estimation systems~\citep{Sundvall2006} to find inconsistencies. 
Moreover, the recent work of \citet{Nobili2018} learns a model to predict failures for pointcloud alignments.
\citet{Akai2019} infers a failure type based on the distribution of the residual error.
The work of \citet{Zhang2016a}, proposes to analyze the structure of the constraints using the eigenvalues to derive a degeneracy factor.

We approach the problem differently by taking into account the underlying graph structure, precisely its spectral properties.
Thus, making our approach independent of the employed sensor system and enabling us to evaluate the discrepancies at multiple scales to be more precise when resolving the spurious estimations.
\subsection{Graph Signal Processing}
Spectral graph theory is an active research area and has gained popularity in the past years in the context of robotics. Spectral graph theory approaches have been proposed for robotic mapping~\citep{Brunskill2007}, planning~\citep{Indelman2018}, and more recently, in combination with graph neural networks for various robotic tasks~\citep{Chandra2020,Moon2020}. 
In general, graphs are irregular structures and are capable of modeling large, complex, and distributed problems~\citep{Mateos2018}, \textit{e.g.} ~\citet{Egilmez2014} proposes an anomaly detection for spatial proximity of graph nodes using spectral graph filtering. Furthermore, graph signal processing aims at applying signal processing techniques on graph structures, thus allowing the use of existing concepts such as the Laplacian operator~\citep{Sandryhaila2014} and multi-scale analysis~\citep{Hammond2011,Hammond2019}. Similarly, ~\citet{Donnat2018} aims to learn a multi-scale structural embedding using graph wavelets by treating the wavelet coefficients as a probability distribution. A good introduction and overview of graph signal processing are presented in ~\citet{Ortega2017}. 

Our approach also performs a structural analysis of graph signals to detect discrepancies between the onboard and server graphs.
Using localized graph wavelets in the graph domain, our approach directly compares the trajectories at different scales to estimate the severity of the inconsistency of the individual pose graphs.

%% file: 03_preliminaries.tex
\section{Preliminaries}
\label{sec:preliniaries}
This section introduces the fundamental and necessary concepts for analyzing and comparing graph structures in the graph spectral domain. 
We first introduce the underlying methods to use graphs for modeling complex problems.
Next, the analysis of harmonic signals in the Euclidean and graph domain are covered. 
\subsection{Fundamental Graph Theory Review}
In this work, we exploit the graph structure that serves as the primary foundation for most modern SLAM backends~\citep{Cadena2016}.
In particular, we extract the pose information of the factor graphs, \textit{i.e.,} disregarding any other sort of constraints to visual landmarks, GPS sensors, etc.
Thus, we consider in this work, weighted undirected graphs $\mathcal{G}=(\mathcal{E},\mathcal{V},w)$ consisting of a set of nodes $\mathcal{V}$ with cardinality $N$, edges $\mathcal{E}$ and weights $w: \mathcal{E}\mapsto\mathbb{R}^{+}$ denoting how strong two nodes are connected with each other.

A graph $\mathcal{G}$ is uniquely described by $\mathcal{E}$, $\mathcal{V}$ and $w$ in the form of a weighted symmetric adjacency matrix $\mathbf{A}\in\mathbb{R}^{N\times{}N}$ with $\mathbf{A}_{n,m} > 0$ if two nodes $n$ and $m$ are connected.
The weight can be chosen freely, such as, for instance, the spatial proximity, or the number of co-observed landmarks between nodes, but ought to measure the relationship between the nodes.
Another fundamental construct in graph theory, is the degree matrix $\mathbf{D}$, defined as a diagonal matrix with entries $\mathbf{D}_{n,n}=\sum_{n'} \mathbf{A}_{n,n'}$ where $n'$ are all incident nodes of $n$.

Finally, signals in traditional signal processing are often expressed as functions over time, such as $x(t):\mathbb{R}\to\mathbb{R}$, mapping a scalar value to each discrete point in time $t$. 
In a similar vein, signals on graphs are defined as $\gs{x}(n):\mathcal{V}\to\mathbb{R}$, associating a scalar value to each node $n$ in the graph.
While a traditional signal $x(t)$ changes over time, a signal defined on a graph $\gs{x}(n)$ alters between the nodes in the graph, leading to certain variations within the signal. 
Analyzing these signal variations and, consequently, their trends can lead to a more fundamental understanding of the signal's nature and is generally termed spectral or frequency analysis.

\subsection{Euclidean and Graph Spectral Analysis Review}
Integral transformations such as the Fourier transform project a signal onto a Hilbert space enabling the analysis of the characteristic properties of that signal. 
The data in the projected space is a compact representation denoting the information that is more or less prevalent in the input data.
However, different aspects, such as the duration, bandwidth, and discretization, need to be carefully considered for a correct analysis of band-limited signals and to avoid unfavorable effects such as leakage and Gibbs ringing.

The Fourier transform is a fundamental tool for various applications in many different fields ranging from image processing to data science~\citep{Rao2010}.
Specifically, the Fourier transform expresses a signal $x(t)$ as a sum of basis functions for which a typical choice is the complex wave: $\cos{}\left(2\pi{}ft\right) + j\cdot\sin{}\left(2\pi{}ft\right) = \exp{}\left(2j\pi{}ft\right)$.
Thus, the traditional Fourier transform is the inner product of a time-dependent signal $x(t)$ with the harmonic oscillation, \textit{i.e.}
\begin{equation}
    X(f)=\langle{}x(t), \exp\left(2j\pi{}ft\right)\rangle_t=\int_{-\infty}^{\infty} x(t)\exp\left(-2j\pi{}ft\right) dt,
\end{equation}
where $X(f)$ is the spectrum of $x(t)$.
Most interestingly, the basis function $\exp{}\left(2j\pi{}ft\right)$ plays an important role w.r.t. the one-dimensional Laplace operator $\Delta$ in $\mathbb{R}$, \textit{i.e.}
\begin{equation}\label{eq:fgspv2:laplace}
\begin{aligned}
    \Delta\, \exp{}\left(2j\pi{}ft\right) &= \frac{\partial{}^2}{\partial{}t^2} \exp{}\left(2j\pi{}ft\right)\\
    &= 2j\pi{}f \frac{\partial{}}{\partial{}t} \exp{}\left(2j\pi{}ft\right)\\
    &= -(2\pi{}f)^2 \exp{}\left(2j\pi{}ft\right).
\end{aligned}
\end{equation}
Based on the Helmholtz equation, \textit{i.e.} $-\Delta{}x=\lambda{}x$, the initial term $(2\pi{}f)^2$ corresponds to the eigenvalue while the later term $\exp{}\left(2j\pi{}ft\right)$ denotes the eigenfunction.

The Laplace operator is a linear operation that expresses the divergence of a given function's gradient.
Likewise, a similar intuition can be applied for deriving a Laplace operator for irregular graph domains.
In concrete, the Laplace operator $\Delta$ is replaced by the Laplacian matrix $\mat{\mathcal{L}}$~\citep{Ricaud2019a} measuring the variations within graphs.
Generally, for an undirected graph with the two matrices $\mathbf{A}$ and $\mathbf{D}$,  the Laplacian matrix can be obtained by 
\begin{equation}\label{eq:fgspv2:graph_laplace}
    \mathcal{L}=\mat{D}-\mat{A},
\end{equation}
which is, by definition, a symmetric and positive semidefinite matrix.
Hence, it can be decomposed into its eigenvalues $\Lambda{}$ and eigenvectors $\mat{U}$, \textit{i.e.} $\mathcal{L}=\mat{U}\Lambda{}\mat{U}^\top$.
Based on the findings in Eq.~\eqref{eq:fgspv2:laplace}, the eigenvector $\mat{U}$ can be used to define the Fourier basis functions and, therefore, also to derive a Fourier transformation of graph signals. 
As a matter of fact, other constructions of the graph Laplacian $\mathcal{L}$ such as the \textit{normalized graph Laplacian} are also possible and are widely used in literature as different graph Fourier bases. 
We refer the interested reader to the work of~\cite{Shuman2016a} for a discussion on differences between the bases. 

In particular, the graph Fourier transform of a graph function $x$ is given by the expansion of $x$ with the eigenfunctions $u$ of the graph Laplacian $\mathcal{L}$ (\textit{cf.} Eq.~\ref{eq:fgspv2:graph_laplace}), \textit{i.e.}
\begin{equation}\label{eq:fgspv2:gft}
    \gs{X}(\lambda_l)=\langle{}\gs{x}(n), u_l(n)\rangle{}_n=\sum_n\gs{x}(n)u_l^*(n),
\end{equation}
where $\lambda_l$ is the $l$-th non-negative eigenvalue of $\mathcal{L}$ and $u_l^*$ is the complex conjugate of $u_l$.
The graph Fourier transform of a graph signal $\gs{x}$ is then given by $\gs{X} = \mat{U}^\top \gs{x}$.
Moreover, the eigenvalues $\Lambda$ are real values, and thus can be ordered and correspond to the graph frequencies, allowing a similar intuition as for traditional frequency analysis.
Consequently, most of a graph signals energy is preserved in the lower bands of $\Lambda$, and higher bands correspond to high oscillating frequencies.

The construction of the graph Fourier transform using Eq.~\ref{eq:fgspv2:graph_laplace} and Eq.~\ref{eq:fgspv2:gft} implies that all the spectral properties are given by the connection of the nodes. 
Thus, the modeling of the relationship between the individual nodes constitutes a crucial component of the system. 

%% file: 04_materials_and_methods.tex
\section{Collaborative Multi-Robot Mapping} 
\label{sec:method}
This section details the proposed method, for which an overview is presented in Figure~\ref{pics:fgspv2:overview}.
Overall, the aim is to identify graph nodes with high drift that will lead to large errors and correct them with only a few constraints. 
The proposed approach comprises the following core components: (i) Onboard localization and mapping, (ii) Mapping server at the base station, and (iii) Pose graph comparison and correction. 
\begin{figure}[!t]
  \centering
   \includegraphics[width=0.46\textwidth, trim={0.0cm, 0.1cm, 0.0cm, 0cm}, clip]{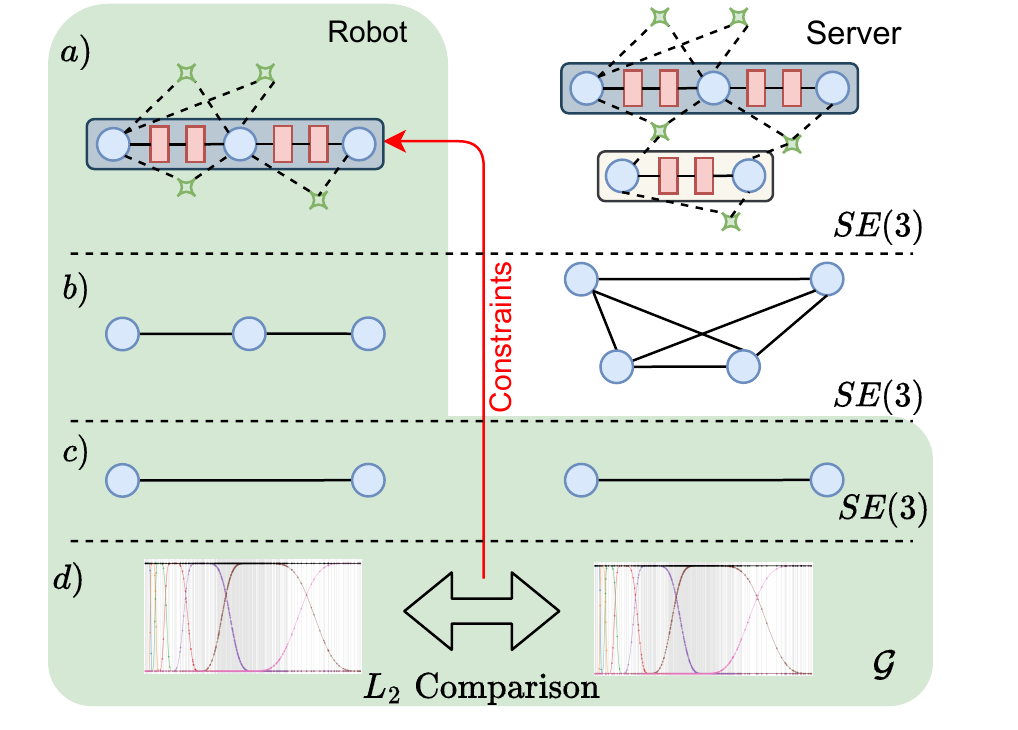}
   \caption{Different employed graphs: a) Onboard visual-inertial graph, which will be incorporated into the global server map. b) Graph abstractions, including Kron reduction. c) Synchronized graphs. d) Multiscale spectral graph analysis. With the results from (d) the system queries poses in $SE(3)$ to generate constraints.}
   \label{pics:fgspv2:method:graphs}
\end{figure}
\subsection{Centralized mapping and Localization}\label{subsec:fgspv2:onboard}
Each robot performs onboard mapping and localization to provide an odometry estimate of its current position as well as to create a local map of the environment. 
Particularly, we don't anticipate that the onboard system must perform any loop closure detection or re-localization methods to reduce their drift.
These methods typically require large computational capacities and might, in real-world deployments, inhibit crucial processes like control, odometry, and navigation.
Hence, in our setup, each robot solely focuses on estimating the relative motion and building an onboard map without optimizing it.
The onboard map and pose estimates are then incrementally sent to a centralized mapping server for additional processing.

More specifically, the mapping server takes the role of a high-performance computing resource and acts as a centralized communications hub for all the robots. 
Hence, the mapping server constitutes the core component of our collaborative mapping approach.
In particular, all incoming robot maps are accumulated, merged into a single collaborative multi-robot map, loop closed, and globally optimized.
Consequently, all the computational-intense multi-robot operations are delegated to the central mapping server while the robots only perform the initial map building.

Since the onboard maps are sent incrementally, the mapping server needs to ensure continuous operation on the global multi-robot map, \textit{i.e.,} it does not run a set of operations once but rather constantly iterates over the global multi-robot map and, thereby, considering the most recent data.

Furthermore, at the end of each iteration, the mapping server constructs a global multi-robot graph from the optimized pose information and broadcasts it on the network.  
Notably, the sent global graph does not contain any sort of constraints such as IMU factors, visual terms, or LiDAR terms, but rather only comprises the positional and rotational information of the nodes.

Each time a robot receives a global graph, it replaces it with the previously received graph until the next update arrives.
Thus, the comparison of the onboard estimation and the globally optimized solution of the server happens at the individual robots using the pose information in $SE(3)$.

\subsubsection{Global Multi-Robot Graph.}
The global graph sent to the robots encapsulates the global knowledge of the environment in a compact representation and contains crucial information such as the last known positions of all robots. 
The global graph is built by defining representative nodes for each incoming submap.
The representative nodes can be freely chosen but ought to reflect the robot trajectories to some degree. 
We found that keyframing heuristics such as minimum distance and rotation between consecutive nodes works well in our experiments. 

Moreover, since the global map is continuously optimized, the graph is not immediately built after every operation, but only a reference to each node is maintained.
When an update is triggered after the global optimization, the \textit{graph monitor} (\textit{cf.} Figure~\ref{pics:fgspv2:overview}) retrieves the latest estimate of the multi-robot pose graph, builds the graph, and sends it to all robots. 

A radius search is performed around each vertex in the graph, and the weight of adjacent nodes is calculated using a squared exponential function where the weight decreases with increasing distance $\Delta$, \textit{i.e.}
\begin{equation}\label{eq:fgspv2:distance_weight}
    w_D(n,m)=\exp\left(-\frac{\Delta(n,m)}{2\sigma^2}\right)
\end{equation}
The distance function $\Delta(n,m)$ measures how much the nodes $n$ and $m$ are related to each other. 
Possible realizations of $\Delta(n,m)$ include measures of spatial proximity or viewpoint coherence, such as distance metrics using the global position $\mathbf{p}\in\mathbb{R}^3$ or global orientation $\mat{R}\in{}SO(3)$, among others.
Although less generic, high-level information, such as the number of co-observed visual landmarks and semantic objects, can also be incorporated into $\Delta(n,m)$.

In our previous work~\citep{Bernreiter2022}, we employed an abstraction from the factor graph, the so-called proxy graph, that only consisted of the positional information in $\mathbb{R}^3$ of the nodes in the factor graph.
Since a drift in the onboard estimation or a degenerate state will always be recognized as a difference in the positions of the estimates. 
However, this does not accurately capture rotational drifts since the positional difference will manifest only on the consecutive nodes.

Therefore, in this paper, we employ a combination of spatial proximity and viewpoint coherence using a distance metric in $SE(3)$~\citep{Barfoot2017}. 
Specifically, between two poses $\mat{T}_n,\mat{T}_m\in{}SE(3)$ we compute
\begin{equation}\label{eq:fgspv2:se3_dist}
\Delta(n,m)=\sqrt{\langle{}\xi_{n,m}^\wedge,\xi_{n,m}^\wedge\rangle}=\sqrt{-\mathrm{tr}\left(\xi_{n,m}^{\wedge}\mat{M}\xi_{n,m}^{\wedge^\top}\right)}
\end{equation}
where $\mat{M}$ is a set of weights for rotation and translation and ${}^\wedge$: $\mathbb{R}^6\to\mathbb{R}^{4\times4}$.
The distance metric $\xi_{n,m}\in{}\mathfrak{se}(3).$ is defined as $\xi_{n,m} = \mathrm{ln}\left(\mat{T}_n^{-1}\mat{T}_m\right)^{\vee}$ with $^\vee{}$: $\mathbb{R}^{4\times4}\to\mathbb{R}^6$ and $\mathrm{ln}$: $SE(3)\to\mathfrak{se}(3)$.

It is important to note that other realizations are possible distance metrics, \textit{e.g.,} in $\mathbb{R}^3$ using the Euclidean distance or in $SO(3)$ using the distance between rotations~\citep{trefethen1997numerical}, \textit{i.e.}
\begin{equation}\label{eq:fgspv2:method:rot_diff}
    \Delta_r(n,m)=\mathrm{tr}\,\mat{R}_m\mat{R}_n^\top{},    
\end{equation}
where $\mat{R}\in{}SO(3)$ and represents the orientation. 

\subsubsection{Graph Reduction.}
Our framework allows for multiple different reduction schemes, for which all approaches operate on the server graph $\mathcal{G}_{server}$ at the centralized mapping server.

When the global graph $\mathcal{G}_{server}$ is broadcasted over the network, each robot synchronizes the onboard estimation with the nodes in $\mathcal{G}_{server}$ based on their timestamp (\textit{cf.} Figure~\ref{pics:fgspv2:method:graphs}c).
Since $\mathcal{G}_{server}$ was built based on the odometry estimates, there is a direct mapping between the nodes.

Moreover, it is important to note that the larger the server graph $\mathcal{G}_{server}$ becomes, the higher the computational and bandwidth requirements are.
This makes it undesirable to send the full graph to all robots in time-critical applications.
As a consequence, $\mathcal{G}_{server}$ is reduced using a Kron reduction~\citep{Dorfler2010}, when the server graph reaches a predefined threshold of nodes.
The Kron reduction removes a subset of nodes of the server graph while preserving the spectral properties and the adjacency matrix. 
The removal of the nodes is implemented as a Schur complement of $\mathcal{L}$
\begin{equation}
    \mathcal{L}_{\text{reduced}} = \mathcal{L}_{n,n} - \mathcal{L}_{n,\tilde{n}}\left(\mathcal{L}_{\tilde{n},\tilde{n}}\right)^{-1}\mathcal{L}_{\tilde{n},n},
\end{equation}
where $\tilde{n}$ are the nodes marked for removal.

The choice of which nodes to keep in the reduced graph should rely upon the information content and how prominent a specific node is.
Most interestingly, the magnitude of the eigenvalues of $\mathcal{L}$ denotes how much information each graph frequency contains.
Therefore, the nodes to keep during the Kron reduction are selected according to the largest eigenvalues in $\Lambda$.




As a next step, we perform a graph spectral analysis using $\mathcal{G}_{server}$ to identify drifts in the onboard estimation.

\subsection{Spectral Analysis of Graph Signals}\label{sec:fgspv2:method:spectral}
The analysis of spectral components of band-limited signals is a well-established and widely used technique in engineering and research.
This paper uses the theory of spectral analysis and signal processing defined on graphs.
In contrast to the standard spectral analysis, graphs do not assume any underlying manifold.
Therefore, they are well-suited for many robotic applications where graph structures such as pose graphs play a significant role.

\subsubsection{Graph Comparison.}
After the robot has received a global update, a chronological synchronization is performed, yielding a one-to-one mapping of the global graph $\mathcal{G}_{server}$ and onboard estimates.
Next, the pose information at the nodes of each graph is used to create the functions $f$ and $h$ for the server and robot estimates, respectively. 
In particular, the pose information for each node in the graph is used to compute the relative distance to the origin yielding a unique expression for each node.

Generally, wavelets are well-known to be very efficient and flexible for a variety of different tasks in signal processing problems~\cite{Hammond2019}. 
In traditional wavelet analysis, a signal $x(t)$ is projected onto a scaled ($a$) and shifted ($b$) wavelet $\psi$, \textit{i.e.}
\begin{equation}\label{eq:fgspv2:classical_wt}
    W(a,b)=\langle{}x(t), \psi_{a,b}(t)\rangle_t=\int_{-\infty}^{\infty} \frac{1}{a}\psi^{*}\left(\frac{t-b}{a}\right)x(t)\,dt.
\end{equation}
Using Parseval's theorem, \textit{i.e.} $\langle{}f, g\rangle = \langle{}F, G\rangle$, eq~\eqref{eq:fgspv2:classical_wt} can also be expressed with the Fourier-transformed signal: $W(a,b)=\langle{}X(\omega),\Psi_{a,b}(\omega)\rangle{}_\omega$.
Analogously, the graph wavelet transform can be derived using the graph Fourier transform and a wavelet filter kernel on $\mathcal{L}$.
For more details, we refer the interested reader to the work of~\cite{Hammond2011, Hammond2019}.
By construction, graph wavelets have the property of being localized on the graph~\citep{Tremblay2014, Donnat2018} and, therefore, directly relate to its structural properties.
The realization of a graph wavelet $\psi_{s,n}$ for a scale $s$ and node $n$ is given by 
\begin{align}\label{eq:fgspv2:psi}
    \mat{\psi}_{s,n} &= \mat{U}\mat{G_s}(\Lambda)\mat{U}^\top\delta_n \\
                     &= \mat{U}\mat{G_s}(\Lambda)\Delta_n,
\end{align}
where $\delta_n$ and $\Delta_n$ are a Dirac centered at vertex $n$ in the graph and graph spectral domain, and $\mat{G_s}$ the wavelet filter bank at scale $s$.
In other words, the filter bank $\mat{G_s}$ acts only on the eigenvalues of the graph, \textit{i.e.} $\mat{G_s}(\Lambda)=\text{diag}(g(s\lambda_1), ..., g(s\lambda_N)$ and is multiplied with the graph Fourier-transformed Dirac, followed by an inverse transform $\mathbf{U}$.
Since $\mat{\psi}_{s,n}$ lies in the graph domain, we can compute the wavelet coefficients for a graph signal $f$ using
\begin{equation}\label{eq:fgspv2:wavelet_coeffs}
    \mat{W}_{s,n}=\mat{\psi}_{s,n}^\top{}f.
\end{equation}
It is important to note that the wavelet $psi$ is calculated only for the server graph, while we compute graph signals for both server and onboard estimation.
Since $psi$ does not change as long as the global server graph does not change, we only need to compute Eq.~\eqref{eq:fgspv2:psi} once per version of the global graph.

We employ the Meyer wavelet due to its good localization in the graph and frequency domain.
Generally, larger values for $s$ compress the filter function $g$, while smaller values of $s$ stretch $g$.
Thus, very large scales do not capture drifts precisely anymore.
We found that using less than ten scales provides a good amount of information with different granularity on the inconsistencies.

The graph signal $f$ is localized for each node in the graph and can be chosen arbitrarily but should reflect the characteristics of each node, similar to the edge weights in Eq.~\eqref{eq:fgspv2:distance_weight}.
In this work, we utilize the distance metric defined in Eq.~\eqref{eq:fgspv2:se3_dist} from the local map origin $o$ to every node $n$ in the graph, \textit{i.e.,} $\Delta(n,o)$.

In summary, we compute wavelet coefficients $\mat{W}$ up to scale $s_{\text{max}}$ for the server and onboard graphs, which constitute a feature vector that represents multiscale structural information for each node in the graph.
Moreover, by comparing the coefficients for the server and onboard estimates, our algorithm can efficiently identify structural differences between them.

\subsection{Correcting Onboard Estimation}\label{sec:fgspv2:method:correction}
For a server node $n$ with corresponding onboard node $n'$, the scale-wise distance is computed as:
\begin{equation}\label{eq:fgspv2:dist}
    d_{n,n'}^s = \norm{\mat{W}_{s,n} - \mat{W}_{s,n'}}_2,
\end{equation}
where $\mat{W}_{s,n}$ and $\mat{W}_{s,n'}$ were computed using Eq.~\eqref{eq:fgspv2:wavelet_coeffs} and the graph signals for server and onboard graph, respectively.

Intuitively, since graph wavelets are localized at a specific node $n$ in the graph~\citep{Tremblay2014, Donnat2018}, and large scales compress $g$, the process spreads further into the graph, leading to a description of the larger neighborhood. 
In contrast, since small scales stretch $g$, they yield a description of the closer neighborhood of $n$.
It is also possible to combine multiple scales, \textit{e.g.,} by accumulating the distances in Eq~\eqref{eq:fgspv2:dist} over multiple scales. 

\begin{figure}[!t]
  \centering
   \includegraphics[width=0.48\textwidth, trim={0.0cm, 0.2cm, 0.0cm, 0cm}, clip]{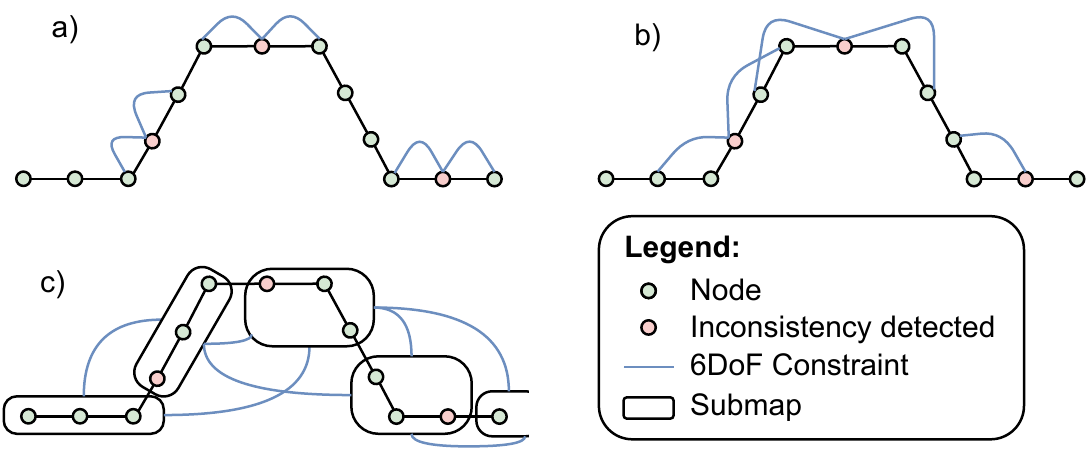}
   \caption{Illustration of three different relative constraint types. Based on the scale of the structural difference, additional constraints are added to correct (a) adjacent, (b) close neighborhood, (c) between submaps, or (d) on degenerate poses.}
   \label{pics:fgspv2:method:constraints}
\end{figure}

Consequently, three separate cases are distinguished, \textit{i.e.}, a large difference in the lower, mid, and higher scales of the coefficients (\textit{cf.} Figure~\ref{pics:fgspv2:method:constraints}a-c). 
In our previous work~\citep{Bernreiter2022}, we introduced thresholds for $d_{n,n'}^s$ to decide when to construct constraints for nodes exceeding it. 
Essentially, the threshold for the small, mid, and large scales needed to be set carefully for the graph and might require some tuning.
In this work, we introduce the notion of only using the top distances for constructing the constraints.
If a small-scale difference is amongst the top distances, a relative constraint is added between the direct neighbors of $n'$. 
Likewise, a corresponding constraint within a multi-hop distance of $n'$ is added for a top mid-scale difference.
In the case of a top large-scale discrepancy, a constraint between the $k$-nearest submaps is added.
It is important to note that the server and robot graph can be expressed in arbitrary and unknown frames since all different types of constraints are relative between nodes.
Additionally, it should be noted that the constraints originate directly from the $SE(3)$ graph that was used for the spectral comparison. 

\subsubsection{Updating the Onboard Graph}
Each robot runs an onboard incremental graph optimization proposed by \cite{Kaess2012} to incorporate the odometry constraints as well as the additional constraints provided by our algorithm.

Since the mapping server will continuously optimize the global multi-robot map and incorporate new inter- and intra-robot loop closures, the constraints can change over time, \textit{i.e.,} existing constraints can be republished with some differences.
Thus, the onboard graph manager module keeps a history of all the incorporated constraints and checks for each incoming constraint whether an already active constraint exists between the corresponding nodes. 
For efficiency, existing constraints are only updated when there is a reasonable difference in translation and rotation to the previous one.
Otherwise, they remain unchanged.

New constraints are not directly added to the onboard graph, but we rather buffer all incoming constraints and apply them at once in a batch. 
This avoids the graph being accessed for every constraint that is identified. 

Additionally, since the global multi-robot graph is modified over time, new constraints are also generated if previously labeled nodes have changed but have not been identified during the current comparison.
This is important since loop closures at the mapping server can significantly alter the global graph.
Accordingly, the newly generated constraints could contradict the already existing constraints in the onboard graph and, therefore, need to be updated.

%% file: 05_experiments.tex
\section{Experiments}
\label{sec:experiments}
We thoroughly evaluate the proposed framework with its different configurations and demonstrate its real-world application using various datasets comprising aerial and legged robots.
First, we validate our approach and compare its performance to the current state-of-the-art methods using the EuRoC~\citep{Burri2016} dataset sequences to simulate multi-robot deployments.
Next, we demonstrate the real-world performance of our framework during a multi-robot autonomous exploration and mapping mission conducted in an underground tunnel system using ANYmal~\citep{ANYmal} legged robots.
Finally, a multi-robot experiment conducted in indoor and outdoor environments demonstrates the localization recovery for an individual agent in case of onboard localization failure.

For all experiments, we use a radius search of $7\,\mathrm{m}$ around the nodes in the global multi-robot map to construct the global graph using Eq.~\eqref{eq:fgspv2:distance_weight}.
The root-mean-square of the absolute trajectory error denoted as RMSE, is used as an evaluation metric for all experiments.
Moreover, unless otherwise stated, our framework is continuously running along with the onboard estimation and performs the proposed spectral analysis every $20\,\mathrm{s}$.
At the end of each comparison, the onboard graph will be provided with the top 15 constraints.

Finally, the mapping server provides an updated multi-robot graph to the robots after performing one cycle of operations.
\subsection{EuRoC Dataset: Validation and Comparison}
We compare the performance of our proposed approach against the current state-of-the-art collaborative mapping frameworks~\citep{Karrer2018, Schmuck2019, Campos2021a} using the Machine Hall (MH) sequences from the EuRoC dataset to evaluate and validate our single- and multi-robot performance.
For each sequence, ROVIO~\citep{Bloesch2017} is used to provide monocular visual-inertial odometry for individual aerial agents.

First, we evaluate the localization performance by comparing the onboard robot estimates, collaborative server estimates, and the proposed approach to the ground truth.
In addition, we also compare current state-of-the-art approaches with results presented in Table~\ref{tab:fgspv2:euroc}.
Since the onboard estimation does not have an initial collaborative result, we omit reporting any error.
Each robot employs a $SE(3)$ distance as in Eq.~\eqref{eq:fgspv2:se3_dist}, unless otherwise stated.
Moreover, each robot does not employ graph reduction measures but utilizes the full graph to infer the constraints.

It can be noted that despite the more significant individual onboard error, the proposed framework still attains the lowest collaborative error. 
Furthermore, by correcting the onboard estimation using our multi-scale spectral approach, the lowest single-robot errors are also achieved, demonstrating the proposed approach's effectiveness in correcting large onboard estimation errors.
\begin{table}[!htb]
    \setlength{\tabcolsep}{3pt}
    \centering
    \begin{tabular}{cccc|c}
        \toprule
        \multicolumn{5}{c}{\textbf{EuRoC Machine Hall - Single and Collaborative}} \\
        \midrule
        \textbf{Method / Seq} & \textbf{MH01} & \textbf{MH02} & \textbf{MH03} & \textbf{MH01-03} \\
        VINS-mono\tablefootnote{Single robot results from~\cite{Qin2018}. Collaborative result from~\cite{Schmuck2019}.} & 0.12\m & 0.12\m & 0.13\m & 0.074\m \\ 
        ORB-SLAM3\tablefootnote{Monocular visual-inertial results from~\cite{Campos2021a}.} & 0.062\m & 0.037\m & 0.046\m & 0.037\m \\ 
        Onboard & 0.21\m & 0.29\m & 0.41\m &  \\ 
        \midrule
        CCM-SLAM\tablefootnote{As reported in~\cite{Schmuck2019}.}  & 0.061\m & 0.081\m & 0.048\m & 0.077\m \\         
        Proposed $\mathbb{R}^3$\tablefootnote{Results from our previous work~\cite{Bernreiter2022}} & 0.029\m & 0.028\m & \textbf{0.033}\m & \textbf{0.025}\m \\ 
        Proposed $SE(3)$ & \textbf{0.026}\m & \textbf{0.027}\m & \textbf{0.033}\m & \textbf{0.025}\m \\ 
        \bottomrule
    \end{tabular}
    \caption{RMSE comparison for the EuRoC dataset. The top part shows the results of single and collaborative approaches, while the bottom row shows the individual corrected results.}
    \label{tab:fgspv2:euroc}
\end{table}
Next, using the experimental setup described in CVI-SLAM~\citep{Karrer2018}, we demonstrate that the proposed approach can facilitate accurate pose estimation for individual robots by providing collaborative corrections, as shown in Table~\ref{tab:fgspv2:euroc:collab}.
\begin{table}[!htb]
    \setlength{\tabcolsep}{5pt}
    \centering
    \begin{tabular}{c|cccc}
        \toprule
        \multicolumn{5}{c}{\textbf{EuRoC Machine Hall - Collaborative Corrections}} \\
        \midrule
        \textbf{Sequences} & \multicolumn{2}{c}{\textbf{CVI-SLAM}} & \multicolumn{2}{c}{\textbf{Proposed}} \\
        & Single & Multi & Single & Multi \\
        MH01 \& MH02 & 0.224\m & 0.139\m & 0.29\m & \textbf{0.027\m} \\ 
        MH02 \& MH03 & 0.295\m & 0.256\m & 0.41\m & \textbf{0.033\m} \\
        MH04 \& MH05 & 0.412\m & 0.34\m & 0.62\m & \textbf{0.085\m} \\
        \bottomrule
    \end{tabular}
    \caption{Onboard pose RMSE after adding constraints from the centralized server for different dataset combinations.}
    \label{tab:fgspv2:euroc:collab}
\end{table}
\begin{table*}[!htb]
    \centering
    \begin{tabular}{c|ccccc}
        \toprule
        \multicolumn{6}{c}{\textbf{EuRoC Machine Hall - Corrections with Different Reduction Levels (RMSE / Nodes)}} \\
        \midrule
        \textbf{Sequences} & \textbf{Onboard} & \textbf{No Reduction} & \textbf{$20\,\%$ Reduction} & \textbf{$40\,\%$ Reduction} & \textbf{$60\,\%$ Reduction}\\
        MH01 (80\m) & 0.21\m & \textbf{0.026\m} / 192 & {0.040\m} / 153 & {0.041\m} / 115 & {0.048\m} / 76\\ 
        MH02 (73\m) & 0.29\m & \textbf{0.027\m} / 178 & {0.032\m} / 142 & {0.036\m} / 142 & {0.042\m} / 71\\ 
        MH03 (130\m) & 0.41\m & \textbf{0.033\m} / 255 & {0.040\m} / 204 & {0.048\m} / 153 & {0.046\m} / 102 \\ 
        \bottomrule
    \end{tabular}
    \caption{Onboard RMSE after correction w.r.t. different Kron reduction levels. The right value denotes the number of nodes in the server graph after the reduction. With the increasing level of reduction, the efficiency of the algorithm decreases due to having less constrained nodes in the onboard factor graph.}
    \label{tab:fgspv2:euroc:kron}
\end{table*}
\begin{figure*}[!htb]
  \centering
  \includegraphics[width=1.0\textwidth, trim={0.0cm, 0.0cm, 0.0cm, 0.0cm}, clip]{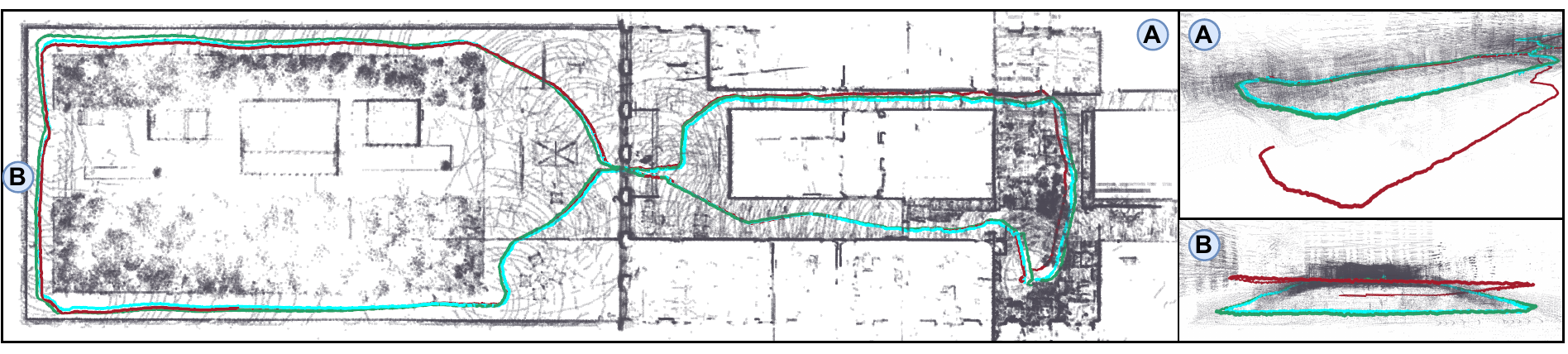}
  \caption{Mapping results for the indoor/outdoor dataset. Cyan denotes the ground-truth trajectory, red the onboard estimation, and green the corrected trajectory. The robot map of ANYmal 1 (red) is misaligned due to the onboard localization failure (A and B), but can be fixed using constraints provided by the centralized mapping server.}
  \label{pics:fgspv2:experiments:h_floor_eval}
\end{figure*}

Finally, we investigate and analyze the effects of applying different reduction levels to the server graph $\mathcal{G}_{server}$.
This leads to a graph with fewer nodes being broadcasted to the individual robots, enabling a smaller network footprint. 
Thus, the server-to-robot communication requirements can be significantly reduced with a sparse relaxation of the dense server graph while still being able to reduce the onboard estimation error.
This is particularly useful for very efficient bandwidth requirements or when the robots are only within connection range for a very short period of time.
Table~\ref{tab:fgspv2:euroc:kron} states the RMSE of the onboard after applying three different levels of reduction to the server graph.

After the reduction, the server graph contains fewer nodes that are sent to the robots. 
Thus, it requires less bandwidth for the transmission to the individual robots.
However, the synchronization of the onboard estimate with $\mathcal{G}_{server}$ leads then also to fewer nodes which results in fewer constraints that can be added to the onboard factor graph, and the error steadily increases with each level.
Nevertheless, even when removing more than $50\%$ of the nodes in the server graph, the error does not significantly increase. 
Consequently, it is possible to use our approach to gain a reasonable improvement with only a few additional constraints in the onboard graph.

\subsection{Analysis of Robotic Drift Recovery}\label{sec:fgspv2:exp:drift}
We conducted an experiment in a particularly challenging environment for LiDAR localization due to the absence of surrounding geometric structure to demonstrate the utility of collaborative mapping towards localization recovery for an individual robot in case of an onboard estimation failure.
Two ANYmal robots were simultaneously deployed in an indoor office environment connected to an outdoor rooftop terrace.
Each robot is equipped with a Velodyne VLP-16 LiDAR and a Sevensense Alphasense visual-inertial sensor. 
Both sensors are synchronized within the onboard computer.

Onboard robot odometry estimation and mapping are performed by CompSLAM~\citep{Khattak2020} and, along with the required visual and pointcloud data, are sent to the mapping server whenever the robots are within the communication range.
The first robot performs a loop indoors while the second robot transitions outdoors through a narrow doorway, navigates a rectangular path, and returns indoors.

\subsubsection{Localization Recovery}
Due to the absence of surrounding structures on the outdoor terrace, the onboard localization drifts significantly, skewing the onboard robot map.
Nevertheless, the collaborative mapping approach is able to generate a consistent map of the environment, as shown in Figure~\ref{pics:fgspv2:experiments:h_floor_eval}, due to its inter- and intra-robot loop closure capabilities. 

We created a ground-truth map using a Leica RTC360 scanner to evaluate the proposed collaborative mapping framework's performance and quantify the effect of the integration of collaborative corrections on individual robot pose accuracy. Ground-truth robot poses were then computed by registering individual robot pointclouds against the ground-truth map following the approach of~\citet{ramezani2020newer}.
\begin{table}[!htb]
    \centering
    \begin{tabular}{ccc|c}
        \toprule
        \multicolumn{4}{c}{\textbf{Indoor/Outdoor Dataset - Ground Truth Evaluation}} \\
        \midrule
        \textbf{Method} & \textbf{RMSE} & \textbf{Time} & \textbf{Server} \\
        ANYmal 1 & 2.22\m & 5.2\,\si{\ms} & \textbf{0.21\m} \\
        Proposed & \textbf{0.30\m} & 5.9\,\si{\ms} &  \\
        \midrule
        ANYmal 2 & 0.25\m  & 3.4\,\si{\ms}  & \textbf{0.14\m} \\
        Proposed & \textbf{0.16\m} & 4.1\,\si{\ms} & \\
        \bottomrule
    \end{tabular}
    \caption{Comparison of the RMSE of the onboard estimation before and after the supplying additional constraints.}
    \label{tab:fgspv2:exp:localization_failure}
\end{table}
Furthermore, the integration of collaborative constraints enables localization recovery for the individual robot leading to a significant reduction in its pose error, as shown in Table~\ref{tab:fgspv2:exp:localization_failure}, when compared to the ground-truth robot trajectory.

The submap constraints are particularly useful for correcting large-scale drifts, while the smaller constraint types performed a local refinement.
Consequently, the submap constraints play a most significant role in the correction and recovery of the onboard drift. 
Hence, in the following, we will investigate the constraint generation process as well as the interrelations between the construction of the graph and the type of constraints.

\subsubsection{Analysis of the Constraint Generation}
In this experiment, we investigate which and where constraints are generated to fix the onboard estimation. 
We additionally compare the performance of different distance metrics and construction approaches of the graphs in comparison to our previous work~\citep{Bernreiter2022}. 

In more detail, we configure our framework to perform the signal and edge weight calculations in different manifolds. 
Specifically, we investigate three different instances of our proposed framework where we use: (i) $\text{L}_2$-Norm in $\mathbb{R}^3$, (ii) using Eq.~\ref{eq:fgspv2:method:rot_diff} in $SO(3)$, and (iii) using Eq.~\eqref{eq:fgspv2:se3_dist} in $SE(3)$.
The specific manifold is not swapped for the graph and signal, \textit{i.e.,} the same manifold is applied for computing the distance for both the edge weights and the signal comparisons.

\begin{figure}[!htb]
  \centering
  \includegraphics[width=0.48\textwidth, trim={0.0cm, 0.0cm, 0.0cm, 0.0cm}, clip]{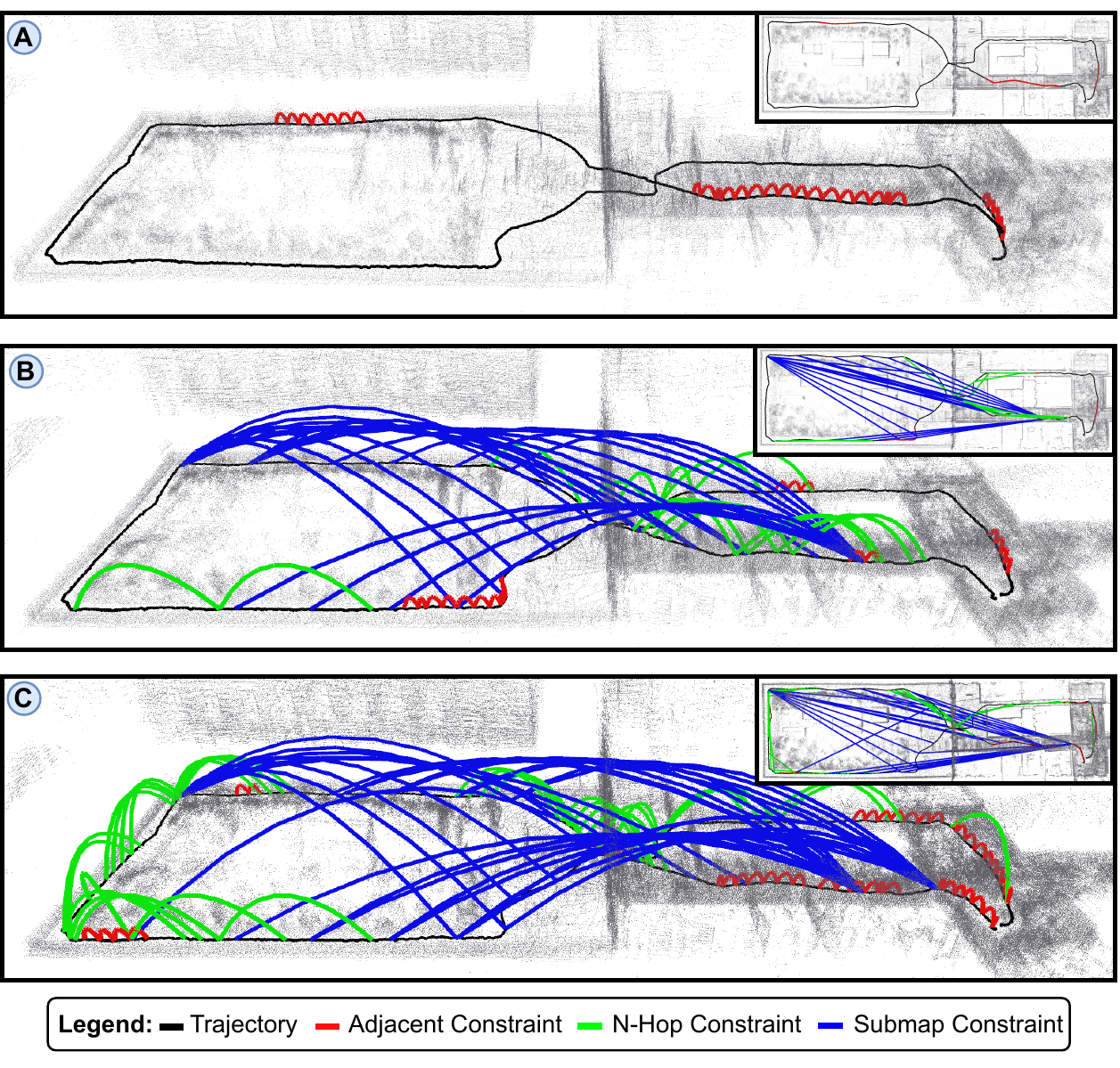}
  \caption{Illustration of the different types of constraints added by using (A) $\mathbb{R}^3$, (B) $SO(3)$ and (C) $SE(3)$. At each comparison, only the top 30 constraints were propagated to the onboard correction node. The top right shows a top-down view of the new constraints.}
  \label{pics:fgspv2:experiments:h_floor_constraints}
\end{figure}

The resulting constraints for each selection strategy are presented in Figure~\ref{pics:fgspv2:experiments:h_floor_constraints}.
It is evident that selecting only the top constraint candidates leads to primarily adjacent constraints for $\mathbb{R}^3$ and n-hop and submap constraints for $SO(3)$.
For the $SE(3)$ strategy, however, it can be seen as a combination of the other strategies, leading to a lower error than for the other approaches.

We present the results in Table~\ref{tab:exp:fgspv2:signal_evaluation} for all the different signal and weight computation strategies.
\begin{table}[!htb]
    \centering
    \begin{tabular}{cccc}
        \toprule
        \multicolumn{4}{c}{\textbf{Indoor/Outdoor Dataset - Constraint Generation}} \\
        \midrule
        \textbf{Constraint Selection} & \textbf{$\mathbb{R}^3$} & \textbf{$SO(3)$} & \textbf{$SE(3)$} \\
        Top 10 constraints & 2.35\m & 1.65\m & \textbf{0.33\m} \\
        Top 30 constraints & 2.05\m & 0.35\m & \textbf{0.32\m} \\
        Top 50 constraints & 1.01\m & 0.33\m & \textbf{0.31\m} \\
        \bottomrule
    \end{tabular}
    \caption{Evaluation of different graph construction and signal generation strategies for the ANYmal 1 indoor/outdoor dataset. }
    \label{tab:exp:fgspv2:signal_evaluation}
\end{table}
Each strategy is separated into \textit{top} 10, 30, and 50 constraints to get a deeper insight into the constraint generation process. 
In this experiment, the onboard estimation is only updated once with the respective \textit{top n} constraints at the end of the run when the server has received and processed all the information to allow for a fair comparison of the individual strategies.
By providing constraints only once with the highest differences, it can be shown how the manifold influences the identification of the drifting nodes. 

Notably, the top 10 constraints for $\mathbb{R}^3$ even increase the error of the trajectory. 
This is due to the fact that solely adjacent constraints were applied to local regions, resulting in a larger error as the drift worsened when assessed globally.
Subsequent evaluations of graphs would then result in more effective constraints.

Furthermore, the results also indicate superior performance of the $SE(3)$ computations. 
Even by only evaluating the top ten constraints, the approach attains a significant improvement to the baseline estimation.
Most importantly, adding more constraints does not account for substantial improvements anymore when using $SE(3)$ computations since the initial result using ten constraints has already identified and corrected the most prominent discrepancies.
\begin{figure*}[!htb]
  \centering
  \includegraphics[width=1.0\textwidth, trim={0.0cm, 0.0cm, 0.0cm, 0.0cm}, clip]{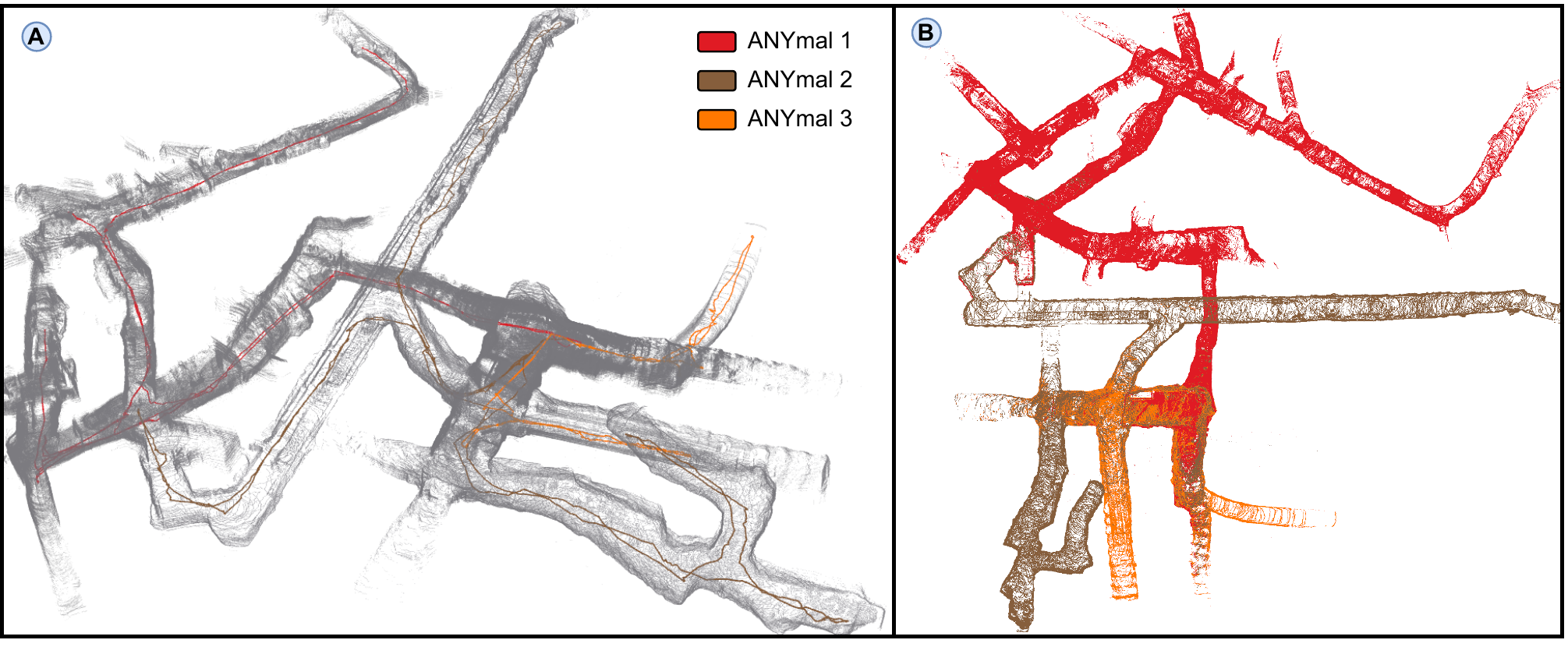}
  \caption{Illustration of the multi-robot map built by our approach. The left image (A) shows the global multi-robot map at the mapping server comprising three individual robot missions. The right image (B) shows a top-down view of the individual maps.}
  \label{pics:fgspv2:exp:hagerbach_hierarchies}
\end{figure*}

\begin{table*}[!htb]
    \centering
    \begin{tabular}{c|ccc|ccc|ccc}
        \toprule
        \multicolumn{10}{c}{\textbf{Underground Tunnel - Ground Truth Evaluation}} \\
        \midrule
        \textbf{Method} & \multicolumn{3}{c}{\textbf{ANYmal 1}} & \multicolumn{3}{c}{\textbf{ANYmal 2}} & \multicolumn{3}{c}{\textbf{ANYmal 3}} \\
        & RMSE & N & NNZ & RMSE & N & NNZ & RMSE & N & NNZ \\
        \midrule        
        Onboard & 1.15\m & 2098368 & 79.2\% & 0.97\m & 1831320 & 79.2\% & 0.43\m & 1070280 & 79.2\,\% \\
        Proposed & {0.14\m} & 5335128 & 50.0\% & {0.10\m} & 7194816 & 34.4\% & 0.07\m & 6140088 & 23.3\,\%   \\
        \bottomrule
    \end{tabular}
    \caption{RMSE comparison for the original onboard, server, and corrected onboard graph. We additionally provide insights into the sparsity of the problem by showing the total number of components (N) and the number of non-zero components (NNZ) per dataset.}
    \label{tab:fgspv2:exp:hagerbach}
\end{table*}
\subsection{Large-Scale Multi-Robot Subterranean Exploration} \label{sec:fgspv2:exp:hagerbach}
We demonstrate the suitability of our approach for complex real-world applications by utilizing it during an autonomous multi-robot exploration~\citep{GBPlanner} and mapping mission conducted at the Hagerbach underground facility in Switzerland.
In this experiment, three ANYmal quadrupedal robots were deployed during an hour-long mission and autonomously navigated distances of roughly $1.2$\,\si{\km}, $1.1$\,\si{\km}, and $600\m$, respectively.
Each robot is equipped with the same sensor payload as described in the previous experiment. 

The collaborative multi-robot map and individual robot maps are shown in Figure~\ref{pics:fgspv2:exp:hagerbach_hierarchies}, with quantitative results presented in Table~\ref{tab:fgspv2:exp:hagerbach}, demonstrating that the collaborative mapping approach greatly benefits from the feedback constraints by significantly reducing the onboard pose estimation error.

In this experiment, we configured our approach to provide as many constraints as possible to achieve the lowest possible onboard trajectory error. 
It is important to note that there is a tradeoff between the incorporation of the constraints into the onboard graph and the resulting improvement it yields.
Table~\ref{tab:fgspv2:exp:hagerbach} also lists the number of components in the optimization problems along with a number of non-zero (NNZ) components. 

Specifically, when incorporating more constraints in the onboard graph, the optimization has to deal with more factors in it, \textit{i.e.}, the problem becomes less sparse as there are more interrelations between the individual factors. 
Hence, when using an incremental solver (iSAM2), as it is running on each robot, it might happen that a large portion of the newly added factors are factors between nodes that are significantly far away in the state matrix, invalidating the locality assumption of the solver along with all its performance optimizations.
As a consequence, the optimization needs to perform more frequently expensive operations and can take a longer time to incorporate the constraints of our approach into the optimization problem.



\subsection{Degeneracy Detection of the State Estimation}
\label{sec:fgspv2:degeneracy}
Next, we investigate the application of our approach in the case of a degenerate condition in the onboard state estimation. 
In particular, each robot employs a LiDAR-based state estimation~\citep{Khattak2020}. 

The experiment took place in a subterranean cave in Switzerland and comprised two legged ANYmal robots and one flying tricopter~\citep{Tranzatto2022}.
Similar to the other experiments, all robots deploy the same sensor payload.
The flying tricopter as well as one of the ANYmal robots, explore the cave environment without any issues. 
However, one of the ANYmal robots enters a long narrow tunnel where its onboard estimation system degenerates for a short period, causing shifted maps in the onboard mapping system. 
This is particularly critical as a safe exploration of the cave cannot be guaranteed any longer for this robot, and it needs to be grounded. 

\begin{figure*}[!htb]
  \centering
  \includegraphics[width=0.88\textwidth, trim={0.0cm, 0.0cm, 0.0cm, 0.0cm}, clip]{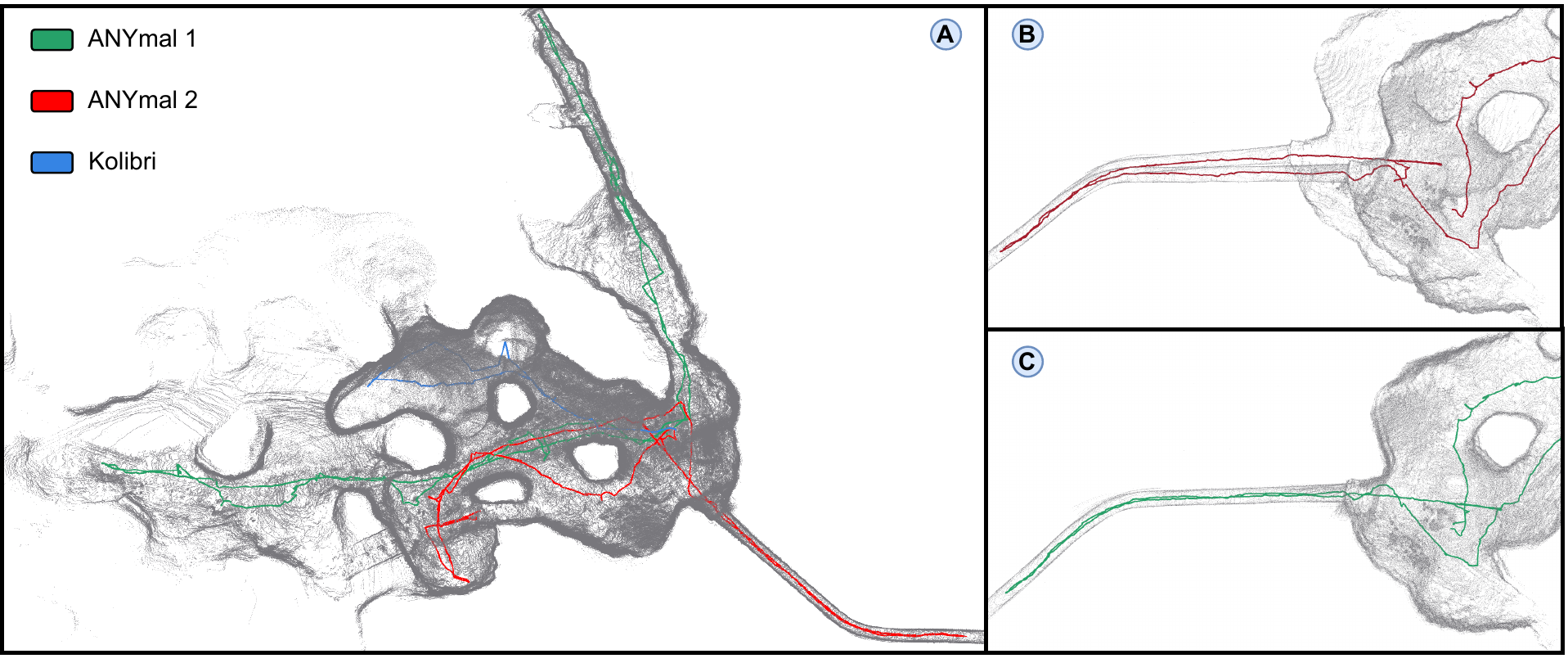}
  \caption{Illustration of the global multi-robot map with corrected robot trajectories (A). The onboard estimation of ANYmal 2 became degenerate during the exploration of a long tunnel (B). The multi-modal mapping server was able to overcome the degeneracy and, by providing additional constraints, fixed the onboard estimation (C).}
  \label{pics:fgspv2:exp:seemuehle}
\end{figure*}

Since our approach is not directly applicable to the detection of degenerate states, we investigate whether, by providing additional constraints, the onboard estimation can recover from it.
Our approach employed a constraint generation strategy of the top 15 constraints and can seamlessly recover from the incorrect state by only using a few constraints.
The global multi-robot map comprises all three robots and, along with individually corrected robot trajectories, is shown in Figure~\ref{pics:fgspv2:exp:seemuehle} together with the degenerate and corrected onboard estimation.

The degeneracy affects the relative position estimation in the direction of the tunnel, \textit{i.e.,} the robot is stuck until the environment becomes unique enough again.
Hence, the degeneracy is recognizable by the fact that the map is estimated shorter in length on the way back than on the way in.
In contrast, a drift in the estimation would have been roughly the same length but still shifted.

Although the multi-modal mapping server requires at least a few optimization cycles to repair the broken map, our approach can still recover the onboard map reasonably. 
Hence, enabling the robot to continue the exploration of the underground environment.

\begin{figure}[!htb]
  \centering
  \includegraphics[width=0.48\textwidth, trim={0.0cm, 0.0cm, 0.0cm, 0.0cm}, clip]{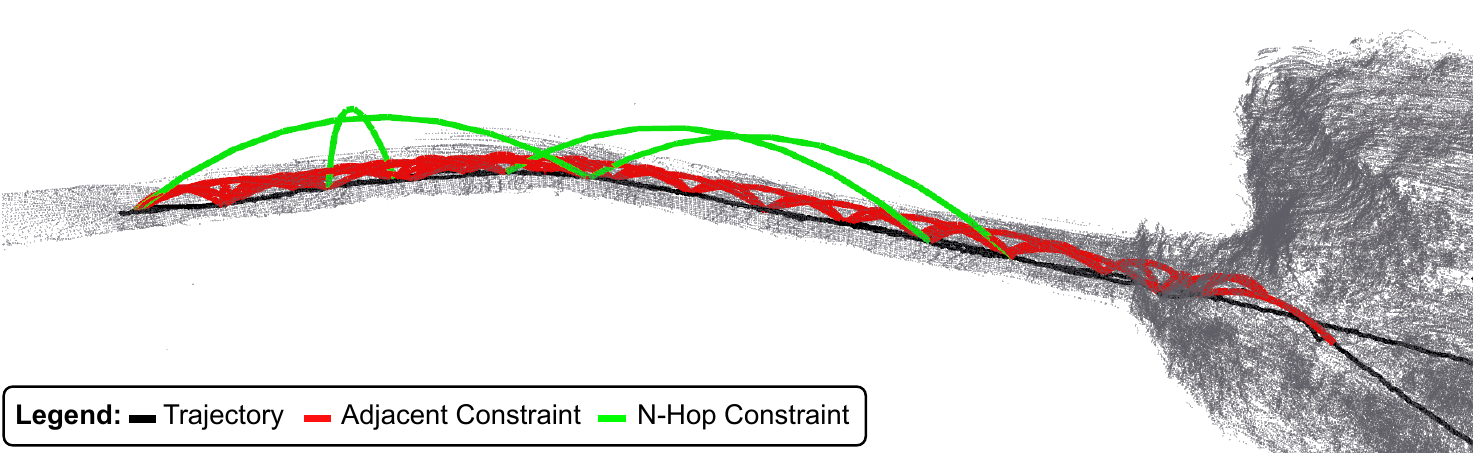}
  \caption{Illustration of the employed constraints constructed by our proposed approach at the degenerate region.}
  \label{pics:fgspv2:exp:seemuehle_constraints}
\end{figure}

In addition, Figure~\ref{pics:fgspv2:exp:seemuehle_constraints} visualizes in detail which constraints our proposed approach sent to the onboard graph.
Most interestingly, the degeneracy in the state estimation of the robot leads to that the prevalence of constructed constraints being between adjacent nodes in the affected area.
Since the degeneracy does not properly estimate the transformation between multiple consecutive steps, the small scales are more predominant than the other scales.

\begin{table*}[!htb]
    \centering
    \begin{tabular}{cccc|c}
        \toprule
        \multicolumn{5}{c}{\textbf{DARPA Subterranean Final Competition}} \\
        \midrule
        \textbf{Robot} & \textbf{Distance}& \textbf{Onboard RMSE} & \textbf{Server RMSE} & \textbf{Corrected RMSE} \\
        ANYmal 1 & 240\m & 0.72\m\,($\pm0.41$\m)  & 0.25\m\,($\pm0.13$\m) & 0.27\m\,($\pm0.14$\m)  \\
        ANYmal 2 & 687\m & 1.29\m\,($\pm0.90$\m) & 0.36\m\,($\pm0.28$\m) & 0.37\m\,($\pm0.41$\m) \\
        ANYmal 3 & 311\m & 0.23\m\,($\pm0.43$\m) & 0.20\m\,($\pm0.34$\m) & 0.22\m\,($\pm0.22$\m) \\
        ANYmal 4 & 500\m & 1.00\m\,($\pm0.71$\m) & 0.24\m\,($\pm0.17$\m) & 0.29\m\,($\pm0.14$\m) \\
        \bottomrule
    \end{tabular}
    \caption{Comparison of the RMSE of the onboard estimation before and after the supplying additional constraints.}
    \label{tab:fgspv2:exp:darpa}
\end{table*}
\subsection{DARPA Subterranean Challenge}
Finally, to show our approach's pertinence to a real-world autonomous multi-robot search and rescue mission, we employ it on the DARPA Subterranean (SubT) dataset of Team CERBERUS~\citep{Tranzatto2022}.
The SubT Challenge was an international robotics competition for fast and autonomous exploration in complex underground environments, where team CERBERUS won the final event. 
Each participating team had to deploy a robotic team that explores an unknown environment reporting the location of specific artifacts in it.

Team CERBERUS' dataset comprises four ANYmal legged robots covering roughly $2\,\mathrm{km}$ of semi-autonomous exploration within a 1-hour mission. 
During the exploration, multiple robots communicated with a central mapping server, providing visual feedback to a human supervisor.
The environment of the final SubT Challenge consisted of three individual regions: i) a tunnel, ii) an urban iii) and a cave environment making it particularly challenging for onboard localization and mapping modules. 

\begin{figure}[!htb]
  \centering
  \includegraphics[width=0.48\textwidth, trim={0.0cm, 0.0cm, 0.0cm, 0.0cm}, clip]{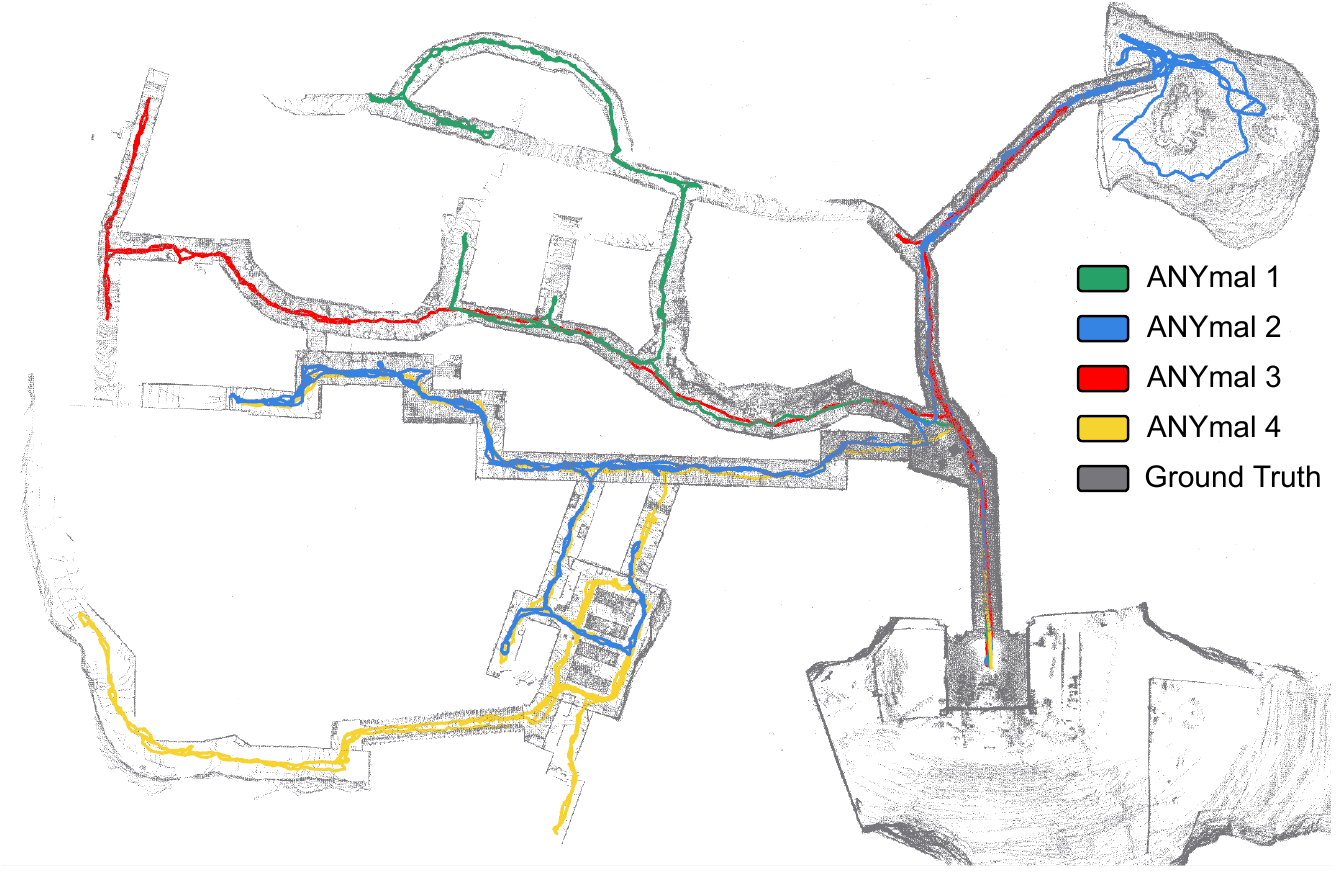}
  \caption{Corrected robot trajectories using our approach with respect to the ground truth map.}
  \label{pics:fgspv2:exp:darpa_overview}
\end{figure}

Although the proposed approach was not deployed during the competition, we show qualitative, post-processed results of the corrected onboard estimations in Figure~\ref{pics:fgspv2:exp:darpa_overview} with respect to the ground truth map.

Moreover, Table~\ref{tab:fgspv2:exp:darpa} shows quantified results for each individual robot and that our approach can significantly improve the onboard estimation modules.
Since none of the robots employed an onboard loop closure detection, each robot accumulated an error over time that could not have been corrected. 
Only the mapping server at the base station runs a multi-robot loop closure detection. 
Hence, by using our approach to construct additional onboard constraints, each robot implicitly incorporates the loop closures from the mapping server and decreases its drift.

It is important to note that Team CERBERUS scored 23 points along with the second-best team and won due to a tie-breaker rule. 
During the exploration of ANYmal 4, it reported an incorrect location of a cell phone (artifact ID: \textit{L22}) artifact for which no score was accounted then.
Cell phone detection was implemented by reporting the current position of the robot when the Bluetooth RSSI values exceeded a specific threshold. 
Thus, the location of the artifact primarily depends on how close the robot walks to the artifact and how large its current drift is.

During the final run, the reported position of artifact \textit{L22} was slightly above the $5\m$ threshold of the ground truth location of the artifact, therefore, was not scored.
The drift of the robot was not significant but just enough to exceed the threshold.  
\begin{figure}[!htb]
  \centering
  \includegraphics[width=0.48\textwidth, trim={0.0cm, 0.0cm, 0.0cm, 0.0cm}, clip]{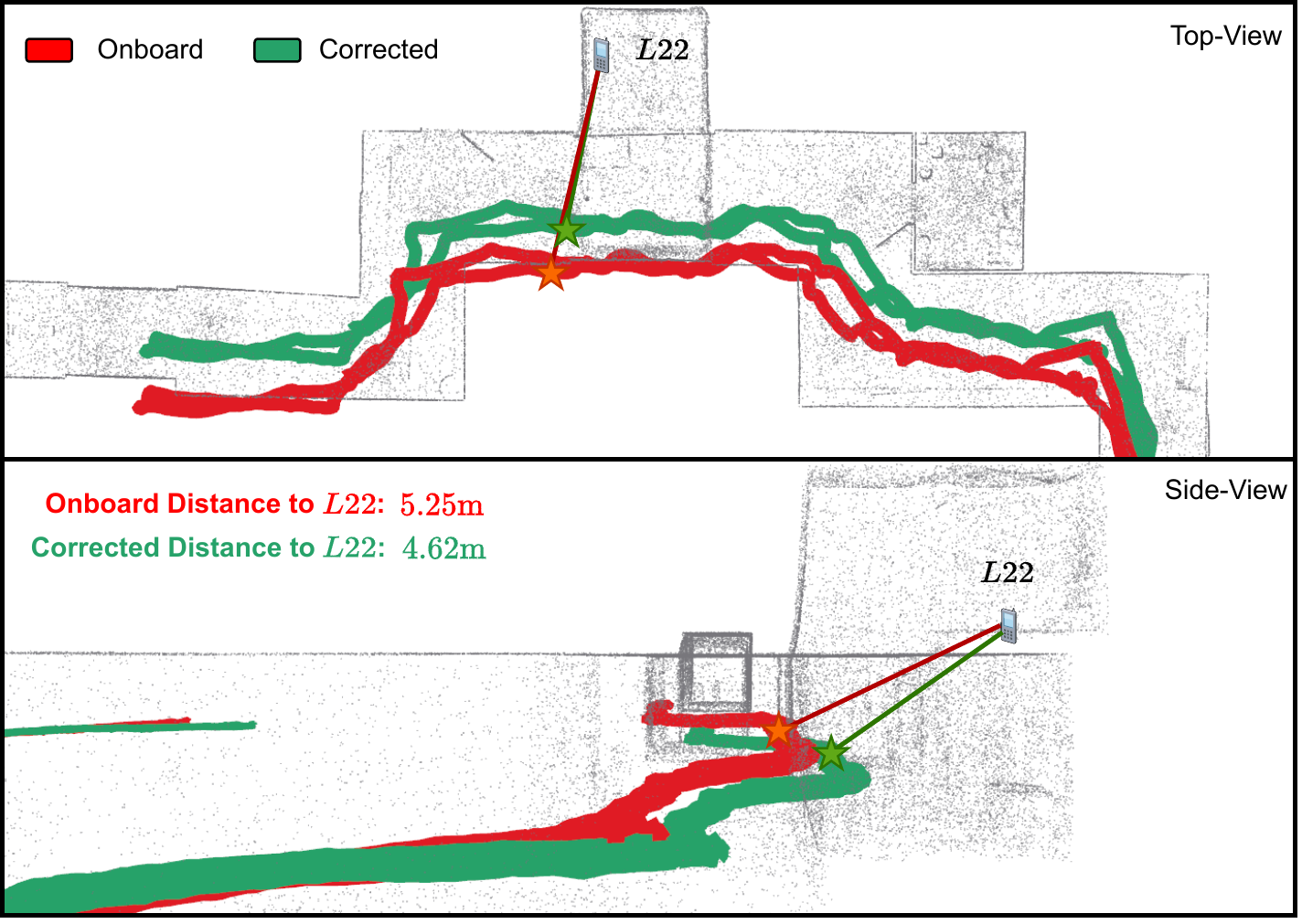}
  \caption{Illustration of the detection of artifact \textit{L22}. The onboard estimation is above the $5\m$ threshold for scoring. Providing additional constraints from our approach leads to a correction that would allow a successful scoring of \textit{L22}.}
  \label{pics:fgspv2:exp:darpa_artifact}
\end{figure}
However, by employing our approach to the post-processed run, the onboard estimation can be greatly improved. 
Figure~\ref{pics:fgspv2:exp:darpa_artifact} illustrates the difference in the reported location of artifact \textit{L22} between the onboard estimation during the final run and the corrected trajectory using our approach. 
As a result, the reported position of the artifact would have been within the $5\m$ threshold and, thus, would have scored.

%% file: 06_conclusions.tex
\section{Conclusions} 
\label{sec:conclusion}
This paper proposed a novel collaborative multi-robot framework for updating the pose graph of individual robots with constraints from a centralized mapping server. 
In this context, we presented a graph-based spectral analysis of the robot and server graphs to identify the underlying structural differences in the onboard estimation.
In particular, our approach computes signals in $SE(3)$ and, along with graph Wavelets, finds nodes in the onboard graph that contradict the globally optimized graph of the mapping server.
Most importantly, by efficiently adding 6-DoF constraints on drifting nodes, minimal additional computation and communication resources are needed, while significantly improving the onboard estimation. 

The presented results using large-scale multi-robot field deployments in challenging environments demonstrate the real-world potential of the proposed approach for both research and industry.

We intend to continue our research in two directions. 
First, we will investigate the construction of graph hierarchies to circumvent the accuracy trade-off for the reduction levels. 
Second, we plan to explore the possibility of keeping the onboard graph sparse using a minimal spanning tree that keeps only a limited set of the most prominent constraints in the onboard graph.